\documentclass{vgtc}

\ifpdf%                                % if we use pdflatex
  \pdfoutput=1\relax                   % create PDFs from pdfLaTeX
  \usepackage{graphicx}                % allow us to embed graphics files
  \DeclareGraphicsExtensions{.pdf,.png,.jpg,.jpeg} % for pdflatex we expect .pdf, .png, or .jpg files
\else%                                 % else we use pure latex
  \usepackage{graphicx}                % allow us to embed graphics files
  \DeclareGraphicsExtensions{.eps}     % for pure latex we expect eps files
\fi%

\graphicspath{{figures/}{pictures/}{images/}{./}} % where to search for the images

\usepackage{microtype}                 % use micro-typography (slightly more compact, better to read)
\PassOptionsToPackage{warn}{textcomp}  % to address font issues with \textrightarrow
\usepackage{textcomp}                  % use better special symbols
\usepackage{mathptmx}                  % use matching math font
\usepackage{times}                     % we use Times as the main font
\usepackage{cite}                      % needed to automatically sort the references
\usepackage{tabu}                      % only used for the table example
\usepackage{booktabs}                  % only used for the table example
\usepackage{caption}

\onlineid{0}

\vgtccategory{Research}
\vgtcinsertpkg

\usepackage{xcolor}
\usepackage{hyperref}
\usepackage{tikz}

\definecolor{orange}{RGB}{215,87,0}
\definecolor{red}{RGB}{220,0,0}

\newcommand{\catcolor}[1]{\textcolor{orange}{#1}}
\newcommand{\dogcolor}[1]{\textcolor{violet}{#1}}
\newcommand{\attackedcolor}[1]{\textcolor{red}{#1}}
\newcommand{\benigncolor}[1]{\textcolor{black}{#1}}
\newcommand{\model}{\textsc{NeuralDivergence}}
\newcommand{\user}{Rachael}
\newcommand{\circledtext}[1]{\raisebox{.5pt}{\textcircled{\raisebox{-.9pt} {#1}}}}
\newcommand{\bigcircledtext}[1]{\raisebox{-.5pt}{\Large\textcircled{\raisebox{-.3pt} {\normalsize{#1}}}}}

\title{NeuralDivergence: Exploring and Understanding Neural Networks by Comparing Activation Distributions}

\author{Haekyu Park\thanks{e-mail: haekyu@gatech.edu}\\ %
        \scriptsize Georgia Institute of Technology %
\and Fred Hohman\thanks{e-mail: fredhohman@gatech.edu}\\ %
     \scriptsize Georgia Institute of Technology %
\and Duen Horng Chau\thanks{e-mail: polo@gatech.edu}\\ %
     \scriptsize Georgia Institute of Technology}
\teaser{
 \centering
 \includegraphics[width=0.98\textwidth]{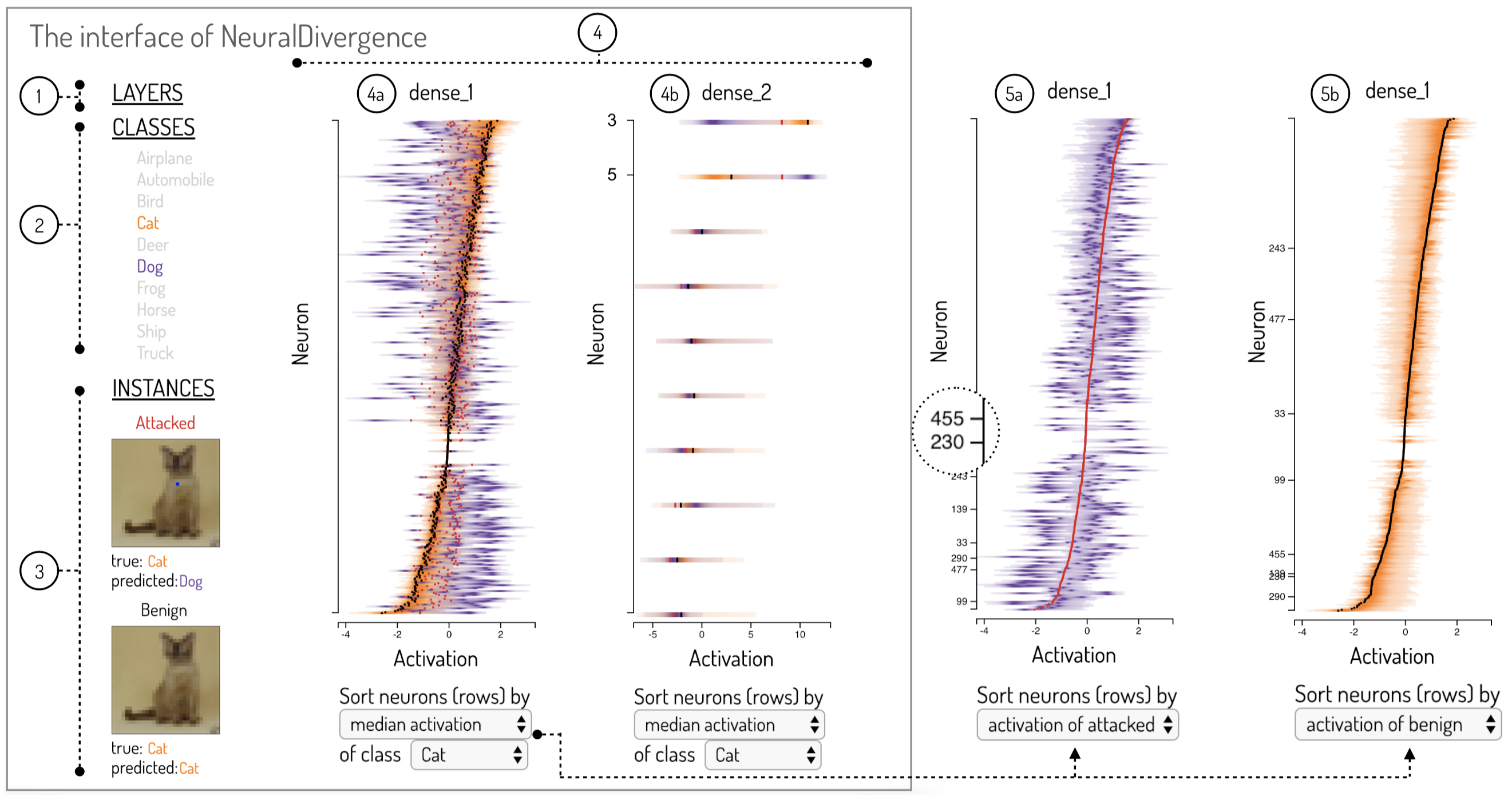}
 \vspace{-0.7em}
 \caption{
    \model{} enables users to explore and understand deep neural networks by comparing aggregated activation distributions across layers, classes, and instances.
    Here, our user \user{} explores how \textit{one-pixel attack} harms a model's prediction.
    She selects:
    \circledtext{1} the network's last two layers (\texttt{dense\_1}, \texttt{dense\_2}) from the \textit{Layers} view; 
    \circledtext{2} the \catcolor{cat} (orange) and \dogcolor{dog} (purple) classes from the \textit{Classes} view;
    and \circledtext{3} a pair of \attackedcolor{attacked} (red) and \benigncolor{benign} (black) cat images from the \textit{Instances} view; one pixel in attacked version has been manipulated to fool the network.
    \circledtext{4} In \textit{Activation} view, each neuron's activation distribution is displayed as a horizontal density bar graph.
    Noting the distribution similarities between  \catcolor{cat} and \dogcolor{dog},
    \bigcircledtext{5a} \user{} wants to discover how activations of the \attackedcolor{manipulated} cat image resemble those of \dogcolor{dogs}', 
    so she sorts the neurons by activations of the attacked input,
    revealing that attacked image, though classified as a dog, is activating the network quite differently from majority of dog images, with large ``neural divergence''.
 }
 \label{fig:teaser}
}

\abstract{
\vspace{-0.5em}
As deep neural networks are increasingly used in solving high-stake problems,
there is a pressing need to understand their internal decision mechanisms.
Visualization has helped address this problem by assisting with interpreting complex deep neural networks.
However, current tools often support only single data instances, or visualize layers in isolation.
We present \model{}, an interactive visualization system
that uses activation distributions as a high-level summary of what a model has learned.
\model{} enables users to interactively summarize and compare activation distributions across layers, classes, and instances (e.g., pairs of adversarial attacked and benign images), 
helping them gain 
better understanding of neural network models.
}

\CCScatlist{
  \CCScatTwelve{Human-centered computing}{Visu\-al\-iza\-tion}{}{};
}

\begin{document}
\maketitle

\section{Introduction}
    Do deep neural networks see the world like humans do?
Given the complex internal structure of neural networks,
this remains a question with no definitive answers. 
In practice, complex models are often used as ``black boxes'', which can be detrimental.
For model developers, they may not know how to fix the model when it fails.
For example, the rich body of research on adversarial machine learning has shown that it is easy to manipulate the pixels of an image in ways that are visually imperceptible to humans, yet would fool a model~\cite{yuan2017adversarial, szegedy2013intriguing}.
To identify the causes of such problems 
and to improve neural network robustness,
it is important to understand how the models operate.

Visualizing neuron activations, internal representations of input data that is transformed into the final prediction, is a useful technique to understand how a model makes predictions~\cite{hohman2018visual, kahng2018activis}.
However, activations are commonly represented as high-dimensional tensors, which are challenging to visualize.
Current tools often support only single data instances, or visualize layers in isolation.

We present \model{}, a system that helps users better understand neural network predictions through interactive summarization and comparison of neuron activation distributions.
\model{}'s major contributions include:

\noindent
\textbf{Summary representation for high-dimensional activations.} 
\model{} compresses high-dimensional activation values of a neuron
into an \textit{activation distribution} representation, 
that can be compactedly visualized, as a horizontal density bar graph (Fig.~\ref{fig:teaser}, at \circledtext{4}). 
These activation distributions provide a high-level summary of the learned representations inside a neural network.

\noindent
\textbf{Interactive, flexible comparison across layers, classes, and instances.}
Popular neural network architectures often consist of many layers.
To gain a more comprehensive understanding of such models' learned representations, it is helpful for a user explore multiple layers at a time, to help discover potential correlations and cross-layer patterns.
Furthermore, an effective way to understand how neural models work is to inspect how they respond to different inputs \cite{kahng2018activis}: 
practitioners often curate ``test cases'' that they are familiar with, so they can spot-check their models; 
also, they often want to perform subset-level analysis (e.g., by considering all images of a class), which works well for large datasets, where instance-by-instance exploration would be too time consuming.
\model{} flexibly supports both instance-level and class-level activation summarization, 
across user-selected layers.

    \vspace{-0.5em}
\section{Illustrative Scenario}
    We provide a scenario to demonstrate how \model{} can help users better understand complex neural network models.
Our user \user{} is a security analyst 
studying the \textit{one-pixel attack}~\cite{su2019one} applied on the VGG16 model~\cite{simonyan2014very}.
When trained on the CIFAR-10 benchmark dataset~\cite{krizhevsky2009learning}, the model attains a 93\% test accuracy.
However, by manipulating just one pixel (see Fig.\ref{fig:teaser}, at \circledtext{3}), the attack can fool the network with great success.
\user{} wants to understand why it works and come up with a countering defense.

\user{} wants to start with concrete examples of successful attacks, 
then generalize her understanding to the whole dataset.
Therefore, she selects a cat image and generates its adversarial version (differs by only one pixel) that would fool the network into mis-classifying it as a dog (Fig.~\ref{fig:teaser}, at \circledtext{3}).
Based on \user{}'s experience working with deep learning models, 
she knows that a model's performance is strongly correlated with its last layers, thus it would be informative to analyze them first.
From the \textit{Layers} view (Fig.~\ref{fig:teaser}, at \circledtext{1}), she selects the last two layers (\texttt{dense\_1}, \texttt{dense\_2}).

While it is helpful to compare the neuron activations of the  \textit{benign} and \attackedcolor{\textit{adversarial}} cat images, 
\user{} expects it would be even more beneficial to compare them with those of \catcolor{\textit{all cat images}} and  \dogcolor{\textit{all dog images}}, 
which may help her discover the similarities of the two classes, and ultimately understand why changing one pixel is sufficient to fool the network.
So, in the \textit{Activation} view (Fig.~\ref{fig:teaser}, at \circledtext{4}), she visualizes them all.
Each neuron's activation distribution is visualized as a binned horizontal bar graph, whose opacity encodes activation densities (a more opaque bin means more images produces that bin's activation values).
As the adversarial cat image is a single image, its activation values are shown as \attackedcolor{\textit{red}} dots in the plots.

Noting the distribution similarities between  \catcolor{\textit{cat}} and \dogcolor{\textit{dog}} (not surprising for both being four-legged animals),
\user{} refines her focus to discover how activations of the \attackedcolor{\textit{manipulated}} cat image resemble those of \dogcolor{\textit{dogs}}'.
So, she visualizes only those two distributions and sorts all neurons by the activation values of the attacked input (Fig.~\ref{fig:teaser}, at \bigcircledtext{5a}).
This reveals that the attacked image, though classified as a dog, is activating the network quite differently from most dog images. 
In other words, the adversarial image (\attackedcolor{\textit{red dots}}) significantly diverges from the ``norm'' of the true class (\dogcolor{\textit{opaque purple regions}}).
She wonders if the amount of ``neural divergence'' between an image and its predicted class could be used to detect possible attacks.
To verify, she visualizes the activations of a benign cat image and that of the cat class (Fig.~\ref{fig:teaser}, at \bigcircledtext{5b}), and sees that the \textit{neural divergence} is small (\textit{black dots} overlap \catcolor{\textit{opaque orange regions}}).
She believes she could use this insight to develop counter defense for the one-pixel attack, and proceeds to perform more testing.

\section{System Design}
    \label{sec:design}
\model{} consists of two main components:
\textit{Selection} views (Fig.~\ref{fig:teaser}, at \circledtext{1}, \circledtext{2}, \circledtext{3}) 
and \textit{Activation} view (Fig.~\ref{fig:teaser}, at \circledtext{4}).
Below, we describe each component.

\textbf{Selection views} allow users to control which subsets of data are included in the neural activation visualization.
These include the \textit{Layers} view for selecting which layers to visualize (Fig.~\ref{fig:teaser}, at \circledtext{1}),
the \textit{Classes} view for selecting which classes to compare (Fig.~\ref{fig:teaser}, at \circledtext{2}),
and the \textit{Instances} view for selecting a pair of attacked and benign images (Fig.~\ref{fig:teaser}, at \circledtext{3}).
Users can select their desired layers, classes, and data instances by the corresponding toggles: displayed items are colored, and hidden items are grayed out.
Classes and instances are each represented by a unique color.

\textbf{Activation view} displays the neural activation distributions.
For each selected layer, the activations are presented in a horizontal bar graph, where the horizontal axis represents the activation magnitude and the vertical axis represents the individual neurons in the layer.
A horizontal bar for each neuron uses opacity to encode the density of the distribution.
Users can interact with \textit{Activation} view in two ways.
First, they can display or hide neurons by clicking the corresponding activation distribution bars (Fig.~\ref{fig:teaser}, at \bigcircledtext{5a}, the magnified area on the vertical axis).
This allows users to keep track of interesting neurons in different conditions.
Second, users can sort the neurons by multiple methods.
These include sorting by the median activation of 
a class (Fig.~\ref{fig:teaser}, at \bigcircledtext{4a} and \bigcircledtext{4b}), 
an attacked image (Fig.~\ref{fig:teaser}, at \bigcircledtext{5a}), and 
a benign image (Fig.~\ref{fig:teaser}, at \bigcircledtext{5b}).
\model{} also supports sorting by the subtraction of activations of two different items.
Users can select the items from the median activation of classes, an attacked image, or a benign image.

Implemented in JavaScript and HTML, \model{} runs in modern web browsers.
D3.js is used to visualize the activation distributions.
A demo is available at \url{http://haekyu.com/neural-divergence/}.

\section{Conclusion}
    We presented \model{}, an interactive system we are developing that visualizes activation distributions to help understand neural networks.
It enables flexible comparisons across various layers, classes, and instances.
Users can explore neural networks by interacting with the system, such as filtering to their desired classes, marking specific neurons, or applying different sorting options.

\acknowledgments{
This work was supported by NSF grants CNS-1704701, TWC-1526254, IIS-1563816, the ISTC-ARSA Intel gift, and a NASA Space Technology Research Fellowship.
}

\bibliographystyle{unsrt}
\bibliography{main.bbl}
\end{document}